\documentclass[english]{lni}
\pdfoutput=1

\IfFileExists{latin1.sty}{\usepackage{latin1}}{\usepackage{isolatin1}}

\usepackage{graphicx}
\usepackage{fancyhdr}
\usepackage{amsmath}
\usepackage{amssymb}
\usepackage{listings} 
\usepackage{changepage} 
\usepackage[figurename=Fig., tablename=Tab., small]{caption}[2008/04/01]

\fancypagestyle{titlepage}{
\fancyhead[RO]{\small A. Br\"omme, C. Busch, A. Dantcheva, C. Rathgeb and A. Uhl (Eds.): BIOSIG 2019, \linebreak Lecture Notes in Informatics (LNI), Gesellschaft f\"ur Informatik, Bonn 2019 \hspace{5pt} \thepage \hspace{0.05cm}} 
\fancyfoot{}}

\renewcommand{\headrulewidth}{0.4pt} 
\author{Clemens Seibold\footnote{Fraunhofer HHI, Berlin, Germany, clemens.seibold@hhi.fraunhofer.de} ,
Anna Hilsmann\footnote{Fraunhofer HHI, Berlin, Germany, anna.hilsmann@hhi.fraunhofer.de} ,
Peter Eisert\footnote{Fraunhofer HHI and Humboldt University Berlin, Berlin, Germany, peter.eisert@hhi.fraunhofer.de}}
\title{Style Your Face Morph and Improve Your Face Morphing Attack Detector}
\begin{document}

\maketitle

\renewcommand{\refname}{References}
\setcounter{footnote}{3} 
\thispagestyle{titlepage}
\pagestyle{fancy}
\fancyhead{} 
\fancyhead[RO]{\small Style Your Face Morph and Improve Your Face Morphing Attack Detector \hspace{5pt} \thepage \hspace{0.05cm}}
\fancyhead[LE]{\hspace{0.05cm}\small \thepage \hspace{5pt} Clemens Seibold, Anna Hilsmann and Peter Eisert}
\fancyfoot{} 
\renewcommand{\headrulewidth}{0.4pt} 

\begin{abstract}
A morphed face image is a synthetically created image that looks so similar to the faces of two subjects that both can use it for verification against a biometric verification system. It can be easily created by aligning and blending face images of the two subjects.
In this paper, we propose a style transfer based method that improves the quality of morphed face images. It counters the image degeneration during the creation of morphed face images caused by blending.
We analyze different state of the art face morphing attack detection systems regarding their performance against our improved morphed face images and other methods that improve the image quality. All detection systems perform significantly worse, when first confronted with our improved morphed face images. Most of them can be enhanced by adding our quality improved morphs to the training data, which further improves the robustness against other means of quality improvement.
\end{abstract}

\begin{keywords}
biometric spoofing, face morphing detection, image quality improvement
\end{keywords}

\section{Introduction}
Ferarra et al.~\cite{Ferrara14} showed that a synthetic face image that looks similar to two different subjects and contains biometric characteristics of both can easily be created with freely-available tools. Both subjects can use this image to verify their identity against a biometric verification system. 
This attack is called face morphing attack. 
It can be performed by a non-rigid alignment and blending of face images of two subjects.
The simplicity of this attack and the fact that an applicant for an official document like a passport or other ID-cards can provide his/her own printed image in most European countries make face morphing attacks a real and dangerous threat to the integrity for biometric verification systems to border control. Therefore, the detection of such attacks is essential for the reliability of biometric verification systems.

In \cite{Lit1} and \cite{Lit2}, an overview on the state of the art of morphing attack detection (MAD) methods is provided.
One important step in the process of creating a morphed face image (morph) is the blending of two aligned images. During this process high frequency details, like wrinkles, scars or pore structures are smoothed or get lost and the resulting image appears dull. Hence, these characteristics can provide evidence for image manipulations and MAD systems are likely to use these characteristics for detection. 

\begin{figure*}[tb]
\begin{center}
\begin{tabular}{ccc}
\includegraphics[width=0.35\textwidth]{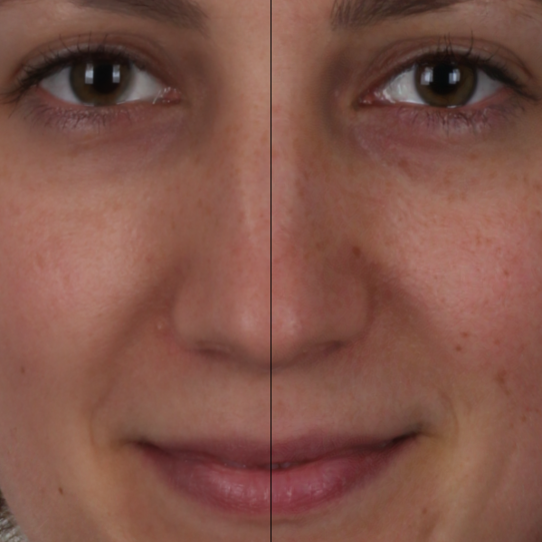}&
\includegraphics[width=0.175\textwidth]{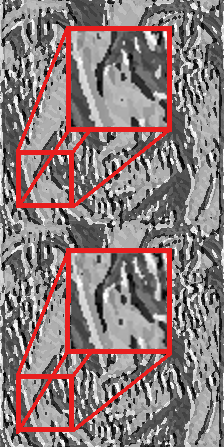}&
\includegraphics[width=0.35\textwidth]{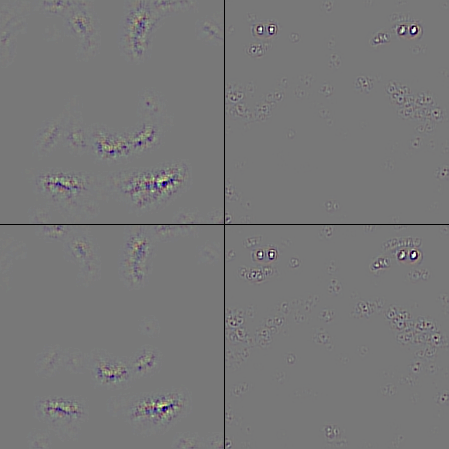}\\
(a) & (b) & (c) 
\end{tabular}
\end{center}
   \caption{(a) simple morph (left half) and improved morph (right half). (b) Binarized Statistical Image Features for a simple morph (top) and an improved morph (bottom). (c) visualizes two feature maps~\cite{Zeiler14} in a later layer with a relative strong difference between simple and improved morph of a DNN that was trained for MAD (top: simple morph, bottom: improved morph). The differences are hardly recognizable in the image as well as in the feature spaces.
}
\label{fig:Teaser}
\end{figure*}

In this paper, we present a style transfer based method that improves the image quality of morphs and counters image degenerating effects caused by the creation process of morphs. We show that being prepared against improved morphs is essential and considering a improvement method can also help to be robust against others. Our improved morphs are more often not detected by MAD systems that were trained without knowledge of these improved morphs, although they only differ slightly compared to simple morphs in image and in feature space, see \texttt{Fig.\ref{fig:Teaser}}. However, most systems can be improved by adding style transfer based improved morphs to the data during training. In addition to robustness against our style transfer based improved morphs, we show that training on these improved morphs can also increase the robustness against other kinds of image quality improvement methods like histogram equalization or sharpening filters. 

{
The key contributions of this paper are:
\setlength{\itemsep}{0pt}
\setlength{\parskip}{0pt}
\setlength{\parsep}{0pt}
\begin{itemize}
\item We propose a Deep Neural Network (DNN) based method that improves the quality of morphs by countering the image degeneration caused by additive blending.
\item We evaluate the vulnerability of several state of the art MAD systems against our improvement method.
\item We show that considering our improved morphs during training can improve the robustness of detection systems also against other image quality improvement methods.
\end{itemize}
}

In contrast to \cite{Hildebrandt17}, \cite{Spreeuwers18} and \cite{Ferrara19} who show that image degeneration operations like adding noise, double-scaling or a print-scan process can worsen a MAD system's performance significantly and need to be considered, we improve the image quality by adapting neural style transfer and study the effects of such improvement on MAD systems.\\  

In the next section, we provide theoretical background on style transfer and our adaptation to improve the quality of morphs. We give a short description of the analyzed MAD systems, our data sources and pre-processing in Section \ref{sec:ExperimentalSetup}. Section \ref{sec:Results} contains our research questions and our experimental results. 

\section{Style Transfer for Morph Enhancement}
\label{sec:StyleTransfer}
\begin{figure*}[!tb]
\begin{center}
 \includegraphics[width=1\linewidth]{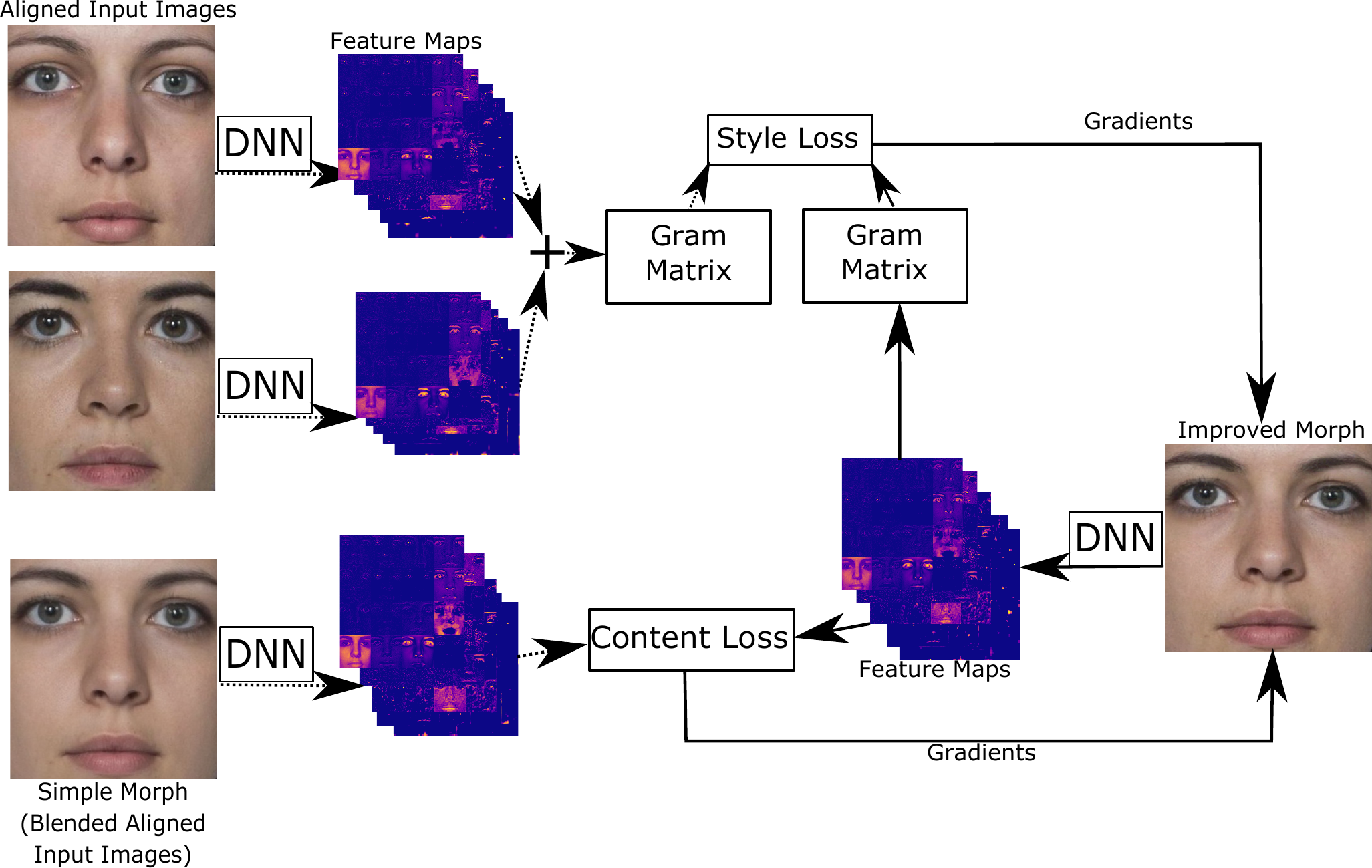}
\end{center}
   \caption{Style transfer for the improvement of morphs.}
\label{fig:StyleTransfer}
\end{figure*}

Gatys et al.~\cite{GatysEB15} proposed a method that changes the style of an image (content image) to the style of another image (style image) while preserving the original content, e.g.~they transformed still images to have the characteristics of famous paintings. The concept of this approach is to transform both input images into a content and a style space and to find an image that is close to the content image in content space and close to the style image in style space. Both spaces are defined by feature maps of a neural network. 
In contrast to Gatys et al., we do not want to transfer an image to a target style, but want the style of our image being similar to the styles of two different face images. For this purpose, we define our target style as average style of both input face images that are used for the creation of a morph, see \texttt{Fig.\ref{fig:StyleTransfer}}.
\paragraph{Mathematical Model for Style Transfer}
The content of an image is defined as vectorized feature maps $F^l_j \in \mathbb{R}^{M_l}$ of selected layers in the neural networks and the style of an image as Gram matrices $G^l \in \mathbb{R}^{N_l \times N_l}$ of $F^l$, with $N_l$ being the number of features maps in layer $l$ and $M_l$ the number of pixels of a feature map in layer $l$. The element $G^l_{i,j}$ of the Gram matrix is the inner product of the vectorized feature map $i$ and $j$ in layer $l$: $G^l_{i,j} = F^l_{i} (F^l_{j})^\mathsf{T}$.
In order to transform the style of an image, we look for an image $I$ that minimizes the loss function
\begin{align*}
&L(I) = \sum_l v_l C(I)_l + \sum_l w_l S(I)_l,
\end{align*}
where $C_l$ and $S_l$ are the content and style loss (the weighted square difference of the content/style of $I$ and the target content/style) of layer $l$ and $v_l$ and $w_l$ are weights for the content and style loss.

We use a VGG-19 network trained for object recognition to calculate the feature maps and minimize $L(I)$ using the gradient based Limited-memory Broyden-Fletcher-Goldfarb-Shanno algorithm with box constraints (L-BFGS-B)~\cite{LBFGSB} to create a new morph with the style characteristics of the original image. 
Style transfer can handle images of any size, since only the convolutional layers of the neural network are used and no fixed size for the feature maps are needed.

\begin{figure*}[tbh]
\begin{center}
\setlength{\tabcolsep}{-1pt}
\begin{tabular}{cccc}
\includegraphics[width=0.24\textwidth]{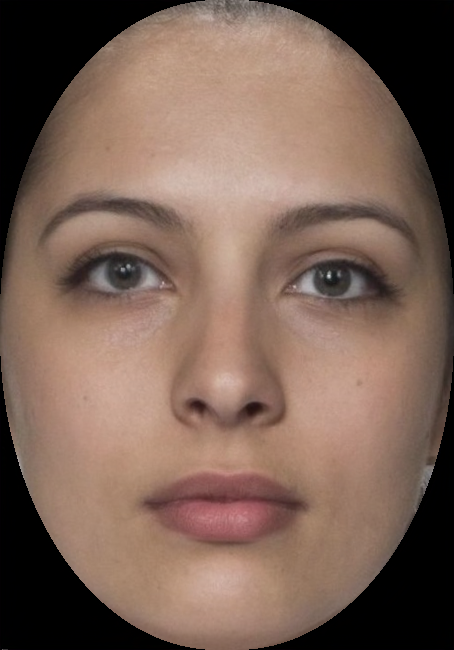} &
\includegraphics[width=0.24\textwidth]{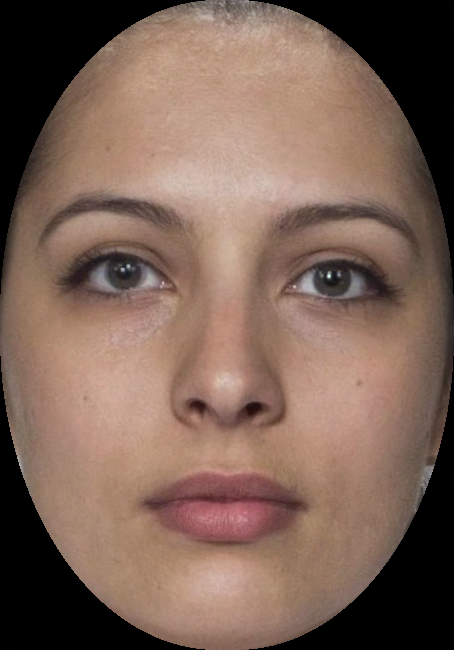} &
\includegraphics[width=0.24\textwidth]{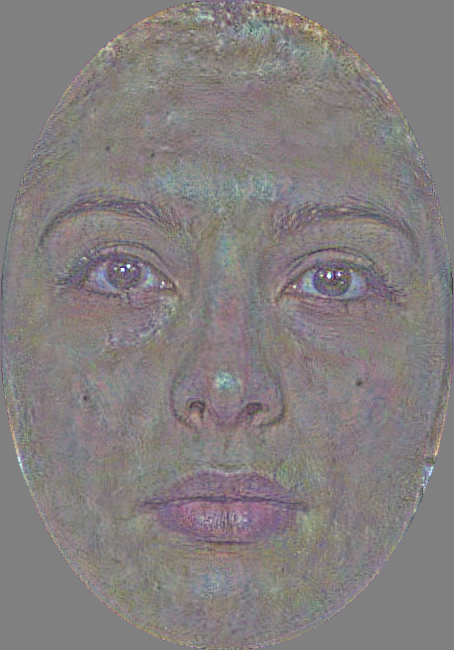}&
\includegraphics[width=0.265\textwidth]{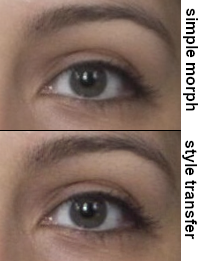}\\
(a) simple morph& (b) style transfer & (c) difference image & (d)\\
\includegraphics[width=0.24\textwidth]{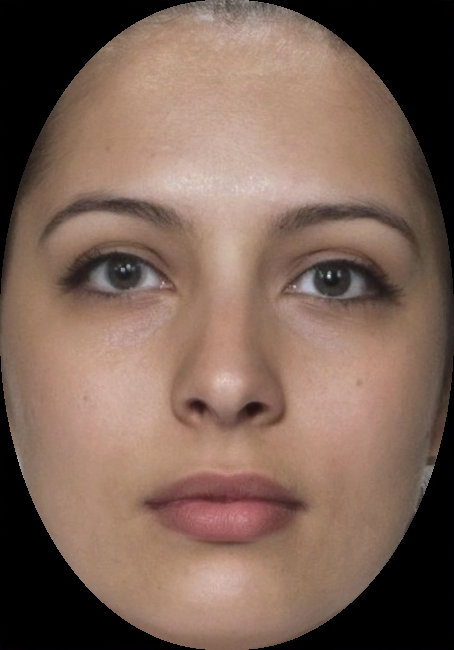} &
\includegraphics[width=0.24\textwidth]{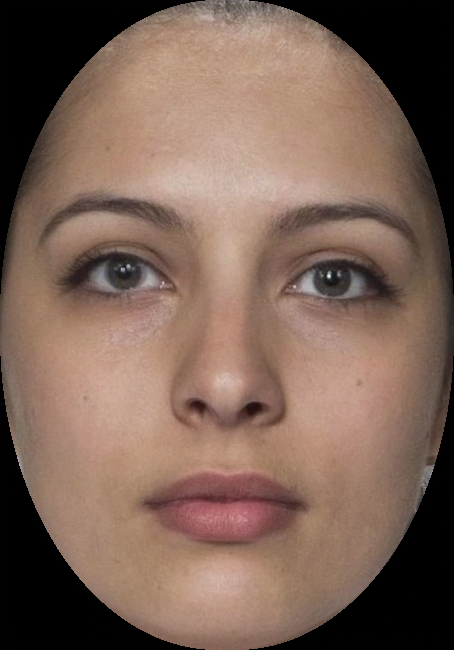} &
\includegraphics[width=0.265\textwidth]{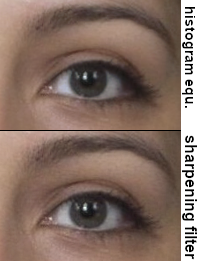}\\
(e) histogram equalization  & (f) sharpening filter & (g)\\
\end{tabular}
\end{center}
 \caption{Examples of different means for the image quality improvement of morphs:
(c) shows difference image between (a) and (b) with enhanced contrast, (d) enlarged part of (a) and (b), (g) enlarged part of (e) and (f)}
\label{fig:ResImprovement}
\end{figure*}

\paragraph{Improving Blended Faces Using Style Transfer}
Before extracting the style from our genuine input images, we align and crop the inner part of the face, see \texttt{Fig.\ref{fig:ResImprovement}a}. The style is extracted as described above for both images and finally averaged to get a target style, see \texttt{Fig.\ref{fig:StyleTransfer}}. The content is provided by cropping the same region of the aligned and blended face image. We initialize the L-BFGS-B algorithm with the simple blended image and use the layers 'conv1\textunderscore2', 'conv2\textunderscore2', 'conv3\textunderscore4', 'conv4\textunderscore4', 'conv5\textunderscore4' for the content representation and 'conv1\textunderscore1', 'conv2\textunderscore1', 'conv3\textunderscore1', 'conv4\textunderscore1', 'conv5\textunderscore1' for the style representation.

\texttt{Fig.\ref{fig:ResImprovement}} shows examples for different means of image quality improvement of morphs.
\texttt{Fig.\ref{fig:ResImprovement}b} shows the style transfer based improved version of the simple morph shown in \texttt{Fig.\ref{fig:ResImprovement}a}. 
The differences between the improved and simple version are visualized in \texttt{Fig.\ref{fig:ResImprovement}c}. Fine structures on the skin and features like moles are enhanced and edges that are smoothed due to the blending, e.g.~in and around the eyes, are reinforced to look sharp again, see \texttt{Fig.\ref{fig:ResImprovement}d}. 
\texttt{Fig.\ref{fig:ResImprovement}e} and \texttt{Fig.\ref{fig:ResImprovement}f} show the results of other means of post-processing that aim to recover the sharpness of the original images. \texttt{Fig.\ref{fig:ResImprovement}e} shows the result of sharpening the image using unsharp masking and \texttt{Fig.\ref{fig:ResImprovement}e} the image after transforming the intensity of the simple morph so that its histogram approximately matches the histogram of one of the input images used for this morph.

\section{Experimental Setup}
\label{sec:ExperimentalSetup}
\paragraph{Morphed Face Image Generation and Improvement}
For the generation of our morphs, we use the all-automatic face morphing pipeline in \cite{Seibold17}. The factor for the geometry warping and the morphing factor for the blending of the images was set to 0.5 for the generation of the morphed images. We apply the improvement of the morphs on the \textit{simple blended} face images before the seamless region cloning is applied (see \cite{Seibold17}).

\paragraph{Face Image Collection and Pre-processing}
We collected face images from different publicly available datasets\footnote{BU-4DFE, Chicago Face Database, FERET, LondonFace Database, PUT, scFace, Utrecht, CyberExtruder Ultimate Face Matching Data Set} and from our internal face databases. In total, we use about 2,000 face images in our experiments, after removing images with bad quality or violating requirements for passports, e.g.~without frontal view. We split these genuine face images into a training dataset including 70\% of all images, test dataset (20\%) and validation dataset (10\%) to avoid overfitting of our detectors based on neural networks.
When selecting pairs for the generation of morphs, we ensure that both subjects have the same gender, are from the same database and are used equally frequent. 

We process all images to have a standardized region and size for the detectors as in \cite{Raghavendra16, Raghavendra17, Seibold18a}. We rotate each image such that the eyes are on a horizontal line and crop the smallest bounding box that includes mouth and eyebrows. Finally, this region is scaled to 224x224 pixels.
\paragraph{Face Morphing Detectors}
In our study, we analyze five different MAD systems. In the following, a short overview on these detectors is provided. 

The image degeneration based MAD system presented in \cite{Kraetzer2017} relies on the number of edge describing features that are detected in an image and the change of the amount of these features after compression.  
The classification is performed using a pruned C4.5 decision tree.

One detector presented in \cite{Raghavendra16} is based on Local Binary Patterns (LBP). The histogram of the 59 uniform local binary pattern is calculated to extract features and a support-vector machine is used for classification.

A second detector proposed in \cite{Raghavendra16} is based on Binarized Statistical Image Features (BSIF) with a filter size of $11\times11$ and a bit length of 12. The histogram of the resulting image with 12-bit depth is calculated and a support-vector machine is employed to obtain a classifier.

Two MAD systems based on deep neural networks \cite{Seibold18a} use the VGG19-architecture and start the training with on object classification pretrained DNNs. One network is directly trained on genuine images vs.~morphed faces (DNN naive), while the other network (DNN MC) is first pretrained on partial morphs.

\section{Results}
\label{sec:Results}
\paragraph{Experimental Goals}
We define three goals to study the quality of our style transfer based morph improvement approach and its effects on MAD systems.\\
\textbf{(G1.1):} We analyze whether the MAD systems can abstract from simple morphs and also detect the improved morphs. We train the five described detection systems on genuine images and simple (not improved) morphs and test them on genuine images, simple morphs and different kinds of improved morphs.\\
\textbf{(G1.2):} We study if the MAD systems can be adapted to robustly detect our style transfer based improved morphs by replacing half of the morphs during the training by these improved morphs.\\
\textbf{(G2):} We analyse the biometric quality of the improved morphs in terms of realistic morph acceptance rate ($rMAR$)~\cite{Hildebrandt17} and morph acceptance rate ($MAR$) using a biometric verification system\footnote{Verilook 10.0/MegaMatcher 10.0 Faces Identification Thechnology Algorithm Demo}. We set the threshold for this system such that we have a FAR of $0.1\%$ (according to the vendor specification) as recommended by FRONTEX for automated border control \cite{FRONTEX15} and a even more strict rate of $0.01\%$. 
\paragraph{Evaluation Metrics}
We consider face morphing attack detection as kind of presentation attack detection (PAD) and use the PAD metrics \textit{bona fide presentation classification error rate} (BPCER) and \textit{attack presentation classification error rate} (APCER), which are defined in ISO/IEC 30107-3 \cite{PAISO}.
\texttt{Tab.\ref{tab:G11}a} and \texttt{Tab.\ref{tab:G12}a} show the performance of the studied MAD systems in BPCER and APCER, separated by the kind of improvement that is applied to the morphs.
In addition to the not post-processed morphs (simple) and our presented improvement method (improved), we use a sharpening filter (sharp), which uses the unsharp masking technique, histogram equalization (HEQU), and histgram equalization after our syle transfer based method (imp.+HEQU) as post-processing steps to improve the quality of our morphs. \texttt{Tab.\ref{tab:G11}b} and \texttt{Tab.\ref{tab:G12}b} show the BPCER at different fixed APCER for the MAD systems that allow an adjustment of the error rates by changing the decision threshold.

\begin{table}[h]
\begin{center}
\footnotesize
\begin{tabular}{|l|cccccc|}
\hline
& & \multicolumn{5}{|c|}{APCER(\%)}\\
Detector & BPCER(\%) & \multicolumn{1}{|c}{simple} & improved & sharp & HEQU & imp. + HEQU\\
\hline
Features~\cite{Kraetzer2017} & 32.6 & 17.3 & 54.6 & 43.6 & 49.7 & 74.7\\
LBP~\cite{Raghavendra16} & 25.4 & 21.1 & 60.3 & 58.5 & 35.1 & 65.7\\
BSIF~\cite{Raghavendra16} & 13.3 & 17.3 & 54.9 & 39.4 & 24.7 & 63.1\\
DNN naive~\cite{Seibold18a} & 1.5 & 1.0 & 30.7 & 3.1 & 32.5 & 72.6\\
DNN complex MC~\cite{Seibold18a} & 1.5 & 0.5 & 27.1 & 2.6 & 29.1 & 62.9\\
\hline
\end{tabular}\\[1.5ex]
a) BPCER and APCER at default threshold of the MAD systems\\[1.5ex]
\begin{tabular}{|l|cccccc|}
\hline
& \multicolumn{6}{|c|}{BPCER(\%) at fixed APCER}\\
Detector & APCER(\%) & \multicolumn{1}{|c}{simple} & improved & sharp & HEQU & imp. + HEQU\\
\hline
LBP~\cite{Raghavendra16} & 10.0 & 40.5 & 82.3 & 69.2 & 55.4 & 83.9\\
 & 5.0 & 54.4 & 90.0 & 82.8 & 67.7 & 90.3\\ 
 & 1.0 & 79.5 & 95.6 & 95.9 & 91.3 & 97.2\\[1.5ex]
BSIF~\cite{Raghavendra16}& 10.0 & 24.1 & 72.6 & 46.9 & 33.9 & 75.4\\
 & 5.0 & 35.9 & 78.5 & 67.2 & 46.7 & 85.1\\ 
 & 1.0 & 78.5 & 88.7 & 89.2 & 74.4 & 93.3\\[1.5ex]
DNN naive~\cite{Seibold18a} & 10.0 & 0.5 & 15.4 & 0.5 & 29.7 & 74.9 \\
 & 5.0 & 0.5 & 26.9 & 1.3 & 45.4 & 83.9\\ 
 & 1.0 & 1.5 & 46.7 & 7.7 & 81.0 & 95.4\\[1.5ex]
DNN complex MC~\cite{Seibold18a} & 10.0 & 0.3 & 7.2 & 0.3 & 26.9 & 61.8\\
 & 5.0 & 0.3 & 16.4 & 0.8 & 51.3 & 84.4\\ 
 & 1.0 & 1.0 & 38.5 & 2.3 & 93.8 & 98.5\\
\hline
\end{tabular}\\[1.5ex]
b) BPCER at different fixed APCER\\
\end{center}
\caption{G1.1 Performance of different MAD systems trained on simple morphs and genuine images}
\label{tab:G11}
\end{table}

\begin{figure}
\begin{center}
\includegraphics[width=0.95\textwidth]{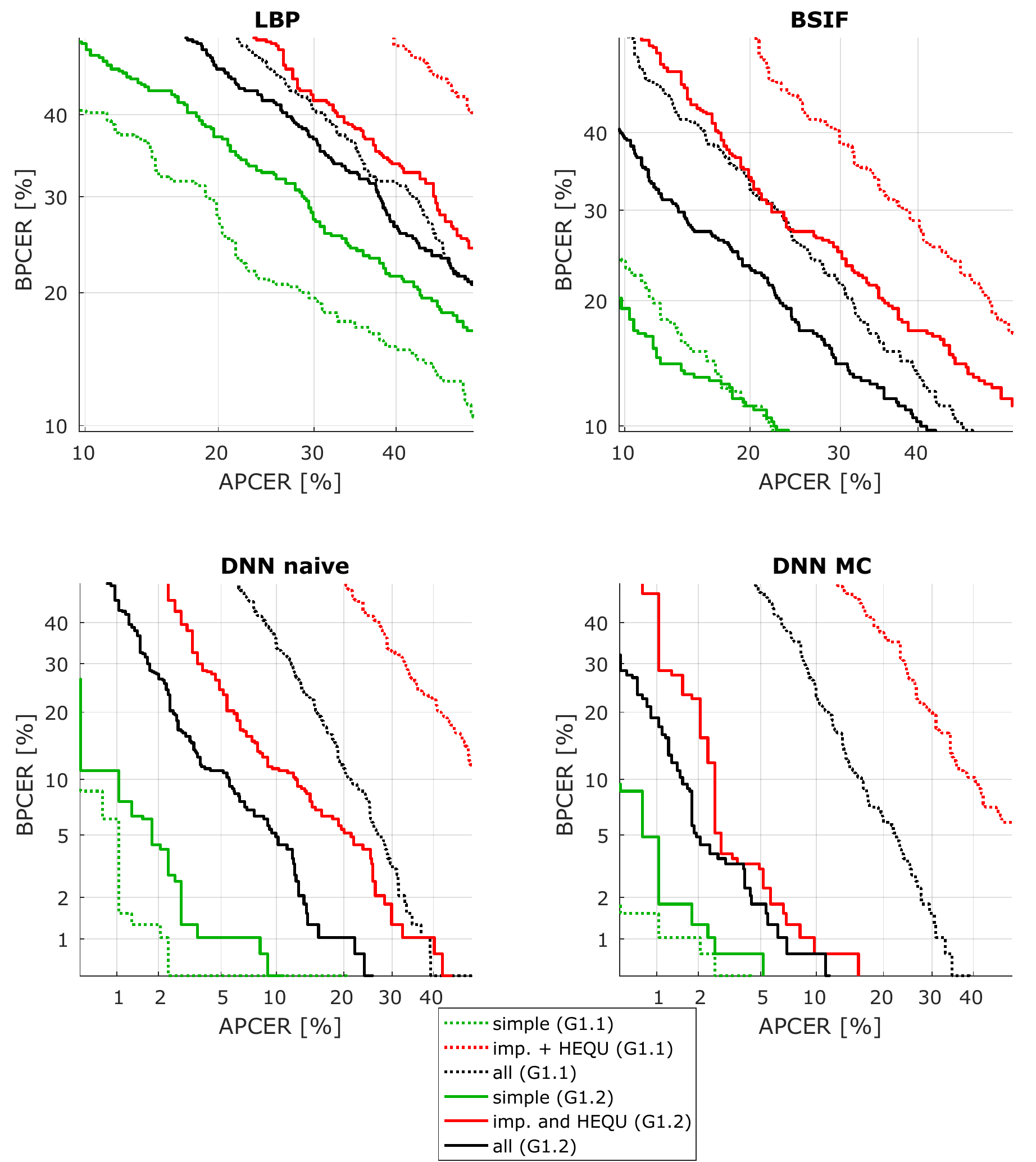}
\end{center}
 \caption{DET curve of the different detectors. Since the Feature-based detector uses a classification tree, no DET curve was calculated.}
\label{fig:DETs}
\end{figure}

All detectors that are trained on genuine images and simple morphs only, performed worse in detecting style transfer based improved morphs and even worse on style transfer based improved and histogram equalized morphs. The APCER increases up to more than 62\% for all detectors. The Detection Error Tradeoff (DET) curves in \texttt{Fig.\ref{fig:DETs}} show that this is not only a matter of threshold of the classifier, but for any given APCER the BPCER is always much worse for style transfer and histogram equalized improved morphs (dashed red line) than for simple morphs (dashed green line).
The MAD systems based on DNNs show the worst absolute and relative loss of performance, while all other methods also perform extremely poor on style transfer based and histogram equalized improved face morphs. 

Including the style transfer based improved morphs in our training data, increases the detection rate of all kinds of improved morphs for all detectors. The detection rate of the style transfer based improved and histogram equalized morphs increases also for nearly all detectors. The detectors adapted differently to this new kind of morph, while the detection systems that are most vulnerable to improvement methods (the DNNs) can best adapt.
The DET curves for the detection of simple morphs (solid green line) of the MAD systems that were trained with also on style transfer based improved morphs are slightly worse for all but the BSIF, but for other means of attacks (solid red for style transfer based and histogram equalized morphs and solid black for all mentioned attacks) far better.

\texttt{Tab.\ref{tab:G2}} shows that our improvement method slightly worsens the biometric quality of the morphed faces, but we still have very high realistic morphing acceptance rates (rMAR)\cite{Hildebrandt17}.

\begin{table}[h]
\begin{center}
\footnotesize
\begin{tabular}{|l|cccccc|}
\hline
& & \multicolumn{5}{|c|}{APCER(\%)}\\
Detector & BPCER(\%) & \multicolumn{1}{|c}{simple} & improved & sharp & HEQU & imp. + HEQU\\
\hline
Features~\cite{Kraetzer2017} & 33.8 & 17.3 & 43.6 & 30.7 & 50.0 & 72.2\\
LBP~\cite{Raghavendra16} & 32.6 & 25.8 & 38.4 & 50.5 & 33.8 & 43.0\\
BSIF~\cite{Raghavendra16} & 17.4 & 10.6 & 31.7 & 34.8 & 19.6 & 38.7\\
DNN naive~\cite{Seibold18a} & 1.5 & 2.8 & 7.2 & 4.4 & 15.7 & 29.9\\
DNN complex MC~\cite{Seibold18a} & 1.8 & 1.8 & 3.4 & 2.1 & 9.3 & 6.4\\
\hline
\end{tabular}\\[1.5ex]
a) BPCER and APCER at default threshold of the MAD systems\\[1.5ex]
\begin{tabular}{|l|cccccc|}
\hline
& \multicolumn{6}{|c|}{BPCER(\%) at fixed APCER}\\
Detector & APCER(\%) & \multicolumn{1}{|c}{simple} & improved & sharp & HEQU & imp. + HEQU\\
\hline
LBP~\cite{Raghavendra16} & 10.0 & 49.2 & 62.8 & 74.4 & 59.0 & 70.5\\
 & 5.0 & 56.2 & 75.4 & 81.5 & 77.7 & 81.0\\ 
 & 1.0 & 84.9 & 87.4 & 89.2 & 89.2 & 89.2\\[1.5ex]
BSIF~\cite{Raghavendra16}& 10.0 & 19.2 & 44.9 & 47.2 & 28.7 & 55.6\\
 & 5.0 & 32.1 & 63.3 & 64.1 & 49.7 & 70.8\\ 
 & 1.0 & 67.4 & 82.6 & 79.0 & 70.0 & 86.2\\[1.5ex]
DNN naive~\cite{Seibold18a} & 10.0 & 0.5 & 1.0 & 1.0 & 6.9 & 11.3 \\
 & 5.0 & 1.0 & 4.4 & 1.3 & 11.8 & 26.4\\ 
 & 1.0 & 7.7 & 11.0 & 10.3 & 50.0 & 60.8\\[1.5ex]
DNN complex MC~\cite{Seibold18a} & 10.0 & 0.0 & 0.0 & 0.0 & 1.3 & 0.8\\
 & 5.0 & 0.8 & 0.8 & 0.8 & 4.4 & 3.3\\ 
 & 1.0 & 1.8 & 3.6 & 3.8 & 21.0 & 28.5\\
\hline
\end{tabular}\\[1.5ex]
b) BPCER at different fixed APCER
\end{center}
\caption{G1.2 Performance of different MAD systems trained on improved morphs, simple morphs and genuine images}
\label{tab:G12}
\end{table}

\begin{table}[h]
\begin{center}
\footnotesize
\begin{tabular}{|l|cccc|}
\hline
Morph type & rMAR1000 & rMAR10000 & MAR1000 & MAR10000\\
\hline
Simple Morphs & 96.0\% & 90.5\% & 98.0\% & 95.3\%\\
Improved Morphs & 93.2\% & 87.0\% & 96.6\% & 93.4\%\\
Sharp & 95.5\% & 90.6\% & 97.8\% & 95.3\%\\
HEQU & 95.5\% & 90.5\% & 97.7\% & 95.2\%\\
Imp.+HEQU & 92.9\% & 86.5\% & 96.4\% & 93.1\%\\
\hline
\end{tabular}
\end{center}
\caption{G2 Biometric evaluation of improved morphs in comparison with simple morphs}
\label{tab:G2}
\end{table}

\section{Summary and Discussion}
We introduced a method that improves the quality of morphs to be a step ahead of the attacker and analyze a broader range of attacks. The performance of different MAD systems drop significantly when they are first confronted with our style transfer based improved morphs. 
After including our improved morphs in the training data, most of the MAD systems get significantly better in detecting them. They are also able to generalize and the detection rate for other means of post-processing that improve the image quality also increases. We achieve the best detection rates for all means of quality improved morphs by the MAD systems based on deep neural networks.

We studied the effects of image quality improvement on MAD systems that are based on handcrafted and learned features and showed that they are sensitive to subtle changes of the image. More sophisticated methods, which are based on physical models like reflection analysis \cite{Seibold18b} or on biometric comparisons \cite{FerraraFM18}, might be more robust against our improvements, since the reflections on the face and the biometric differences are only changed slightly.

Our method for style transfer based improvement of morphs is easy to implement using a deep learning framework and does not need special or expensive resources. Hence, it is a realistic scenario that an attacker would try to improve morphs using style transfer and thus it should be considered in the evaluation of MAD systems.

\section*{Acknowledgment}\label{sec:Acknowledgments}
The work in this paper has been funded in part by the German Federal Ministry of Education and Research (BMBF) through the Research Program ANANAS under Contract No. FKZ: 16KIS0511.

\def\bibindent{1em}


\begin{thebibliography}{123456}

\bibitem[FFM14]{Ferrara14}Ferrara, M.; Franco, A.; Maltoni, D.: The magic passport. In: IEEE International Joint Conference on Biometrics. 2014.

\bibitem[FFM18]{FerraraFM18}Ferrara, M.; Franco, A.; Maltoni, D.: Face Demorphing. IEEE Trans. Information Forensics
and Security, 2018.

\bibitem[FFM19]{Ferrara19}Ferrara, M.; Franco, A.; Maltoni, D.: Face morphing detection in the presence of printing/
scanning and heterogeneous image sources. ArXiv e-prints, January 2019.

\bibitem[FR15]{FRONTEX15}FRONTEX: Best Practice Technical Guidelines for Automated Border Control (ABC)
Systems. 2015.

\bibitem[GEB15]{GatysEB15}Gatys, L. A.; Ecker, A. S.; Bethge, M.: A Neural Algorithm of Artistic Style. CoRR,
abs/1508.06576, 2015.

\bibitem[Hi17]{Hildebrandt17}Hildebrandt, M.; Neubert, T.; Makrushin, A.; Dittmann, J.: Benchmarking face morphing
forgery detection: Application of stirtrace for impact simulation of different processing
steps. In: IWBF 2017. April 2017.


\bibitem[In17]{PAISO}International Organization for Standardization: , ISO/IEC 30107-3:2017 Information
technology - Biometric presentation attack detection - Part 3: Testing and reporting, 2017.

\bibitem[Kr17]{Kraetzer2017}Kraetzer, C.; Makrushin, A.; Neubert, T.; Hildebrandt, M.; Dittmann, J.: Modeling Attacks
on Photo-ID Documents and Applying Media Forensics for the Detection of Facial
Morphing. In: IHMMSec'17. 2017.

\bibitem[MW18]{Lit2}Makrushin, A.; Wolf, A.: An Overview of Recent Advances in Assessing and Mitigating
the Face Morphing Attack. In: 26th European Signal Processing Conference, EUSIPCO
2018, Roma, Italy, September 3-7, 2018. 2018.

\bibitem[Ra17]{Raghavendra17}Raghavendra, R.; Raja, K. B.; Venkatesh, S.; Busch, C.: Transferable Deep-CNN Features
for Detecting Digital and Print-Scanned Morphed Face Images. In: CVPRW. 2017.

\bibitem[RRB16]{Raghavendra16}Raghavendra, R.; Raja, K. B.; Busch, C.: Detecting morphed face images. In: BTAS.
2016.

\bibitem[Sc19]{Lit1}Scherhag, U.; Rathgeb, C.; Merkle, J.; Breithaupt, R.; Busch, C.: Face Recognition Systems
Under Morphing Attacks: A Survey. IEEE Access, 7, 2019.

\bibitem[Se17]{Seibold17}Seibold, C.; Samek, W.; Hilsmann, A.; Eisert, P.: Detection of Face Morphing Attacks by
Deep Learning. In: IWDW 2017, Magdeburg, Germany. 2017.

\bibitem[Se18]{Seibold18a}Seibold, C.; Samek, W.; Hilsmann, A.; Eisert, P.: Accurate and Robust Neural Networks
for Security Related Applications Exampled by Face Morphing Attacks. ArXiv e-prints,
June 2018.

\bibitem[SHE18]{Seibold18b}Seibold, C.; Hilsmann, A.; Eisert, P.: Reflection Analysis for Face Morphing Attack Detection.
In: 26th European Signal Processing Conference (EUSIPCO). 2018.

\bibitem[SSV18]{Spreeuwers18}Spreeuwers, L.; Schils, M.; Veldhuis, R.: Towards Robust Evaluation of Face Morphing
Detection. In: 26th European Signal Processing Conference (EUSIPCO). 2018.

\bibitem[ZF14]{Zeiler14}Zeiler, M. D.; Fergus, R.: Visualizing and Understanding Convolutional Networks. In:
Computer Vision - ECCV 2014. 2014.

\bibitem[Zh97]{LBFGSB}Zhu, C.; Byrd, R. H.; Lu, P.; Nocedal, J.: Algorithm 778: L-BFGS-B: Fortran Subroutines
for Large-scale Bound-constrained Optimization. ACM Trans. Math. Softw., 1997.

\end{thebibliography}
\end{document}